\documentclass[conference]{IEEEtran}
\usepackage{times}
\usepackage{algorithm}
\usepackage{graphicx}
\usepackage{algpseudocode}
\usepackage{amsmath,amssymb}

\usepackage[numbers]{natbib}
\usepackage{multicol}
\usepackage{xcolor}
\usepackage[bookmarks=true,hidelinks]{hyperref}
\usepackage{cuted}
\usepackage{caption}
\newcommand{\projectrepo}{%
\href{https://github.com/ruanjinchen/Continuum-Robot-Modeling-with-Action-Conditioned-Flow-Matching}%
{\textcolor{blue}{TDCR Project Repo}}%
}

\begin{document}

\let\titleold\title
\renewcommand{\title}[1]{\titleold{#1}\newcommand{\thetitle}{#1}}
\def\maketitlesupplementary
   {
   \newpage
       \twocolumn[
        \centering
        \Large
        \textbf{\thetitle}\\
        \vspace{0.5em}Supplementary Material \\
        \vspace{1.0em}
       ] 
   }


\title{Continuum Robot Modeling with Action Conditioned Flow Matching}

\author{%
\authorblockN{Jiong Lin$^{1,*}$, Jinchen Ruan$^{1,*}$, and Hod Lipson$^{1}$}
\authorblockA{$^{1}$Department of Mechanical Engineering, Columbia University, New York, NY, USA\\
$^{*}$Equal contribution.}
}


%

\maketitle

\begin{strip}
    \includegraphics[width=\textwidth]{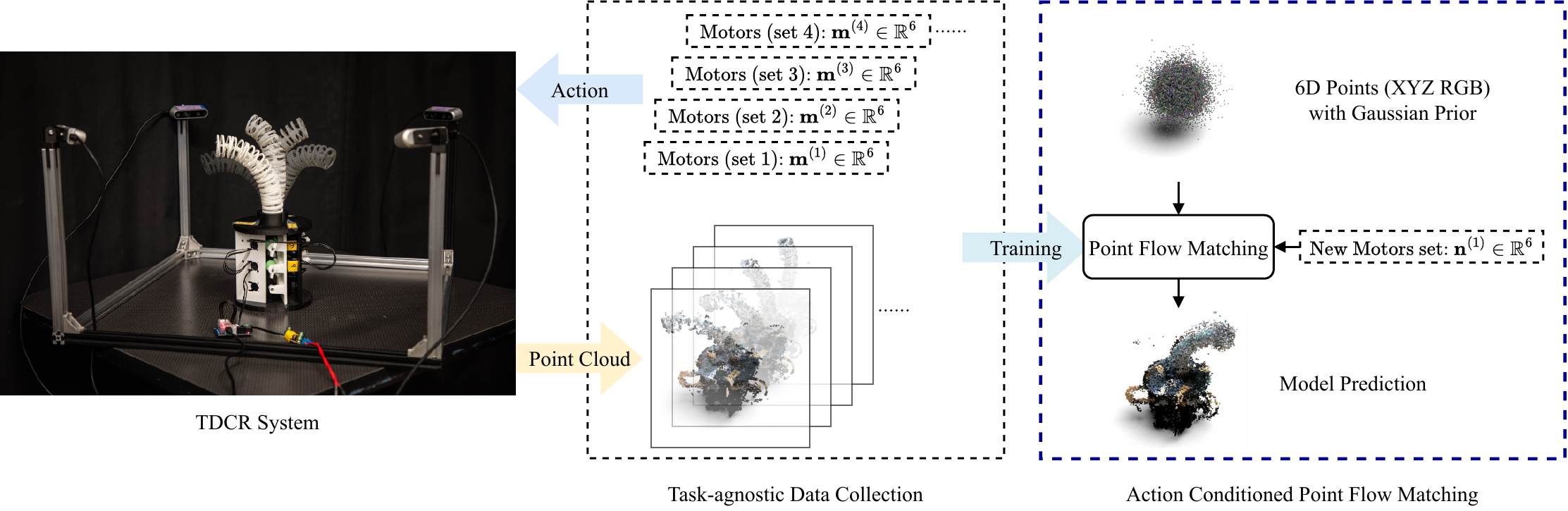}
    \captionof{figure}{
     Overview of our action conditioned point flow matching framework for a tendon driven continuum robot (TDCR) shape prediction. Task independent data samples (motor commands and point clouds) are collected from the real TDCR system, then used to train a flow matching model that predicts robot geometry for new motor inputs.
    }
    \label{fig:teaser}
\end{strip}

\begin{abstract}
Predicting the shape of tendon driven continuum robots (TDCRs) at steady state from actuation remains challenging due to continuous deformation, complex tendon routing, compliance, friction, and fabrication variability. In this paper, we address this problem as kinematic self modeling conditioned on action. We present a lightweight 3D printed TDCR hardware platform and an RGB-D data collection pipeline with multiple cameras, and we learn a point cloud flow matching model that maps motor actuation states to the robot's settled 3D geometry. The model is trained from randomly sampled quasi static configurations and evaluated on test motor commands within the same TDCR design family and actuation range. We compare against prior 3D deformable object and robot self modeling approaches in both MuJoCo simulation and real hardware experiments. Experiments on simulated 2-, 3-, and 5-module TDCRs and real 2- and 3-module robots show improved shape prediction accuracy under CD and EMD metrics. We further show in simulation that the same conditional formulation generalizes to tip payload as a conditioning input, enabling payload conditioned steady-state shape prediction. These results demonstrate a data driven self modeling framework for quasi static TDCR geometry prediction. Project materials are available at: \projectrepo.
\end{abstract}

\IEEEpeerreviewmaketitle

\section{Introduction}

Continuum robots, especially tendon driven continuum robots (TDCRs), offer high dexterity and inherent compliance, making them promising for operation in confined, cluttered, and unstructured environments
\cite{burgner2015survey,walker2013review}.
Unlike rigid link manipulators, TDCRs deform continuously along their bodies, and their observed shapes are affected by tendon routing, backbone compliance, friction, hysteresis, fabrication variability, and external loading.
Classical kinematic representations such as constant curvature and piecewise constant curvature models provide compact and interpretable descriptions of continuum robot shape
\cite{webster2010cc,jones2006kinematics}.
More expressive mechanics based formulations, such as Cosserat rod models, can model distributed deformation, tendon routing, and external loads, but often require careful parameter identification and can be costly to calibrate for a specific physical platform
\cite{rucker2011cosserat,rao2021benchmark,chikhaoui2019comparison}.
These challenges motivate data driven models that learn the robot's shape directly from observations.

In this paper, we study TDCR modeling as \emph{action conditioned quasi static self modeling}.
Given a motor or tendon actuation state, our goal is to predict the robot's settled 3D geometry after the robot reaches a steady configuration.
The learned mapping is induced by the robot's physical deformation behavior, but our target is an observation level forward shape predictor rather than an explicit mechanics simulator.
In particular, we do not attempt to recover material parameters, tendon forces, or transient deformation trajectories.
Our primary setting is free space, quasi static TDCR shape prediction within a given robot design family and actuation range.
We additionally evaluate a payload conditioned extension in simulation by augmenting the condition with a scalar tip payload magnitude, demonstrating that additional quasi static state variables can be incorporated when such data are available.

A key difficulty in TDCR modeling is that the robot body is high dimensional and continuously deformable.
Predicting only a sparse set of backbone states or a small number of geometric parameters may miss shape details that are important for planning, collision checking, or visual servoing.
We therefore represent each robot configuration as a dense point cloud reconstructed from RGB-D observations from multiple cameras.
This representation is permutation invariant, directly compatible with real hardware depth sensing, and can capture full body bending and twisting without imposing a hand designed parametric shape model.

We propose an \emph{action conditioned flow matching} formulation for learning this point cloud shape predictor.
Rather than directly regressing a deterministic mapping from motor commands to point coordinates, the model learns a conditional velocity field in point cloud space that transports samples from a simple prior distribution to the distribution of robot shapes associated with a queried actuation state.
At inference time, the model generates a dense 3D point cloud by integrating the learned flow from the prior to the predicted robot shape.
The integration variable in this generative process is a transport time used by the flow model, not the physical time of robot deformation; the predicted output is the final quasi static shape.

To evaluate the approach, we build a lightweight, configurable 3D printed TDCR hardware platform and a synchronized RGB-D capture system with multiple cameras.
The hardware is instantiated as 2- and 3-module real robots, while our MuJoCo simulation environments include 2-, 3-, and 5-module TDCR variants.
Across simulated and real hardware settings, we compare against representative point based and view based deformable object and robot self modeling methods, including Visual Self Modeling (VSM)~\cite{vsm}, conditional continuous normalizing flows (PointFlow)~\cite{pointflow}, NeRF inspired self simulation (FFKSM)~\cite{nerf_selfsim}, and articulated 3D Gaussian splatting~\cite{gs_selfmodel}.
Our experiments show that action conditioned flow matching achieves lower geometric error in steady state shape prediction, including more complex multi module settings and a simulated payload conditioned setting.

In summary, our main contributions are:
\vspace{-0.2em}
\begin{itemize}
    \item A lightweight, configurable 3D printed TDCR platform and RGB-D capture pipeline with multiple cameras for collecting real hardware self modeling data.
    \item An action conditioned flow matching framework for quasi static TDCR self modeling, which predicts dense settled 3D robot geometry from motor actuation states.
    \item A systematic evaluation in MuJoCo simulation and on real TDCR hardware, including comparisons with point based and view based baselines, multi module TDCR variants, and a payload conditioned simulation extension.
\end{itemize}
\section{Related Work}
\label{sec:related_work}

This paper connects tendon driven continuum robot (TDCR) modeling, robot self modeling, and conditional 3D point cloud generation.
We review prior work with an emphasis on how existing methods represent continuum robot shape, what physical quantities they model, and how our work differs as an action conditioned quasi static shape predictor.

\paragraph{TDCR design and hardware diversity}
TDCRs are a widely used class of continuum manipulators because tendon actuation enables slender robot bodies, remote actuation, and compliant interaction with confined environments
\cite{burgner2015survey,trivedi2008review,walker2013review,rao2021benchmark,russo2023overview}.
However, TDCR designs vary substantially in structural backbone material, tendon routing, module construction, actuation layout, and manufacturing process.
Prior systems have explored helical spring backbones and redundant tendon actuation \cite{li2018design}, helical tendon routing for enlarged workspace and dexterity \cite{starke2017helical}, extensible or follow the leader sections \cite{nguyen2015extensible,amanov2019extensible,neumann2016ftl}, fully actuated or programmable stiffness segments \cite{grassmann2022fas,dewi2024modular}, universal jointed and compressible backbone designs \cite{shentu2024universal,srivastava2023compressible}, open source TDCR benchmarking platform \cite{deutschmann2022opensource}, and monolithic 3D printed TDCR designs \cite{kierbel2025monolithic}.
Beyond tendon driven mechanisms, octopus inspired soft arms and architectured soft structures such as trimmed helicoids further illustrate how geometry and material topology can be used to tune bending, axial stiffness, workspace, and compliance \cite{mazzolai2012octopus,guan2023trimmed}.
This diversity motivates data driven modeling pipelines that can be instantiated on a given robot platform, and cautions against claiming a single morphology independent model for all continuum robots.
Our hardware is therefore used as a configurable TDCR platform for collecting self modeling data, rather than as a proposed universal TDCR mechanical design.

\paragraph{Kinematics, mechanics, and quasi static TDCR modeling}
Classical continuum robot kinematics often uses constant curvature or piecewise constant curvature (PCC) assumptions to obtain compact shape coordinates and tractable forward/inverse kinematics
\cite{webster2010cc,jones2006kinematics}.
These kinematic representations can provide useful geometric descriptions, but they do not by themselves model material stress, tendon force, friction, or external loading.
Mechanics based approaches, including tendon driven manipulator mechanics, Cosserat rod models, geometrically exact steady state formulations, and loading/stiffness analyses, explicitly reason about distributed bending, shear, torsion, tendon routing, actuation, and external loads
\cite{camarillo2008mechanics,rucker2011cosserat,renda2012steady,dalvand2018loading,oliverbutler2019stiffness,armanini2023structured}.
TDCR specific comparisons and benchmarks further show that model accuracy depends strongly on assumptions about curvature, tendon routing, friction, and calibration quality
\cite{rao2021benchmark,chikhaoui2019comparison}.
Recent learning based TDCR kinematic models also address uncertainty or hysteresis in forward kinematics \cite{thompson2024mdn,cho2024hysteresis}.
Our approach is complementary to these analytical and learned kinematic models.
We learn an observation level map from actuation, and optionally payload, to the robot's settled 3D geometry.
The learned map captures the kinematic deformation seen in the data, but it is not a physical based simulator and does not recover material parameters, tendon forces, or transient deformation trajectories.

\paragraph{Shape sensing and learning based state estimation}
Accurate shape information is central to TDCR control, planning, and self modeling.
Embedded sensing methods, such as fiber Bragg grating (FBG) sensors, can provide high rate curvature, shape, or force estimates, but require sensor integration, calibration, and often robot specific modeling assumptions
\cite{roesthuis2016fbg,khan2017fbgforce,ryu2014fbg}.
External sensing with cameras, stereo, or RGB-D sensors avoids embedding sensors into the continuum body, but the reconstructed shape can still be affected by depth accuracy, calibration error, segmentation quality, and view dependent sampling.
Learning based visual methods have been used for soft continuum shape control and inverse kinematics, showing that data driven models can absorb difficult to model geometric and mechanical effects from observations
\cite{almanzor2023static}.
In contrast to methods that estimate a low dimensional curvature state or image space target, we use RGB-D observations from multiple cameras to supervise dense 3D point clouds and learn a forward self model from actuation to full body geometry.

\paragraph{Robot self modeling and neural 3D reconstruction}
Robot self modeling aims to learn an internal representation of a robot's body from interaction or sensory observations, enabling adaptation when the robot morphology or environment is uncertain
\cite{bongard2006selfmodel,cully2015trial}.
Recent visual self modeling methods use modern perception and neural rendering to reconstruct robot geometry and appearance from images.
Visual Self-Modeling (VSM) learns approximate robot morphology from camera observations and masks \cite{vsm}.
NeRF style implicit representations and explicit 3D Gaussian splatting have also been used for robot self simulation and detailed neural reconstruction
\cite{mildenhall2020nerf,kerbl2023gaussiansplatting,nerf_selfsim,gs_selfmodel}.
These view based methods are powerful for appearance modeling and novel view rendering, but their objectives are not always aligned with predicting the full 3D body geometry from a queried actuation state.
Our method instead operates directly in point cloud space, which matches RGB-D supervision and provides a direct geometry representation for shape prediction, collision reasoning, and downstream planning.

\paragraph{Point cloud generative modeling and flow matching}
Point clouds are unordered and irregular, motivating architectures that process point sets directly, such as PointNet/PointNet++, graph-based models, transformers, and point voxel hybrids
\cite{qi2017pointnet,qi2017pointnetpp,wang2019dgcnn,zhao2021pointtransformer,liu2019pvcnn}.
For 3D generation, continuous normalizing flows such as PointFlow model point cloud distributions through invertible continuous time transformations \cite{pointflow}, while diffusion and score based models have enabled high quality generation and completion of point, voxel, and implicit 3D representations
\cite{ho2020ddpm,song2021sde,zhou2021pvd,nichol2022pointe,jun2023shape}.
Flow Matching trains continuous time generative models by regressing velocity fields along probability paths, providing a simulation free objective for learning transport maps
\cite{chen2018neural,grathwohl2019ffjord,lipman2023flowmatching,liu2022rectified}.
In our setting, the continuous time variable is a generative transport parameter, not physical deformation time.
We use flow matching to learn a conditional transport from a simple prior point distribution to the steady state robot shape distribution associated with a motor command.

\paragraph{Positioning of our work}
Classical TDCR models provide interpretable kinematic or mechanics based structure, but they require modeling choices and calibration that can be difficult for fabricated continuum hardware.
Visual self modeling methods reconstruct robot geometry from observations, but they often focus on image or rendering objectives rather than point cloud prediction from actuation.
Our contribution is to combine a real TDCR data collection platform with an action conditioned flow matching model that predicts dense quasi static 3D robot geometry from motor actuation states.
The method is evaluated on simulated and real TDCRs, and we further test a payload conditioned simulation extension by adding a scalar tip payload condition.

\section{Method}
\label{sec:method}

\subsection{Problem Definition and Notation}
\label{subsec:problem}
We formulate TDCR modeling as action conditioned quasi static self modeling.
For each actuation command, the robot is allowed to settle before observation, and the learning target is the final 3D geometry rather than the transient deformation process.
A training dataset contains $K$ settled configurations,
\begin{equation}
\mathcal{D}=\{(X_1^{i}, c^{i})\}_{i=1}^{K},\qquad
X_1^{i}\in\mathbb{R}^{N\times d},
\label{eq:dataset_tdcr}
\end{equation}
where $X_1^{i}$ is a permutation invariant point cloud with $N$ points and $d\in\{3,6\}$ channels (XYZ or XYZ+RGB).
The condition vector $c^{i}$ contains the normalized action variables used to specify the quasi static state.
For the standard TDCR experiments, $c^{i}=\tilde{m}^{i}$ is the normalized motor/actuation vector.
The raw motor vector $m$ denotes tendon command/rest length states in simulation and servo absolute positions on hardware.
For the payload conditioned simulation, $c^{i}=[\tilde{m}^{i}\,\|\,\tilde{p}^{i}]$, where $p^{i}$ is a scalar gravity aligned tip payload magnitude.
For an $S$-module TDCR, the motor dimension is $D=3S$; thus $D=6$, $9$, and $15$ for the 2-, 3-, and 5-module robots, respectively.
Novel motor commands refer to test commands within the same robot morphology and actuation limits.

\begin{figure*}[t]
    \centering
    \includegraphics[width=0.9\textwidth]{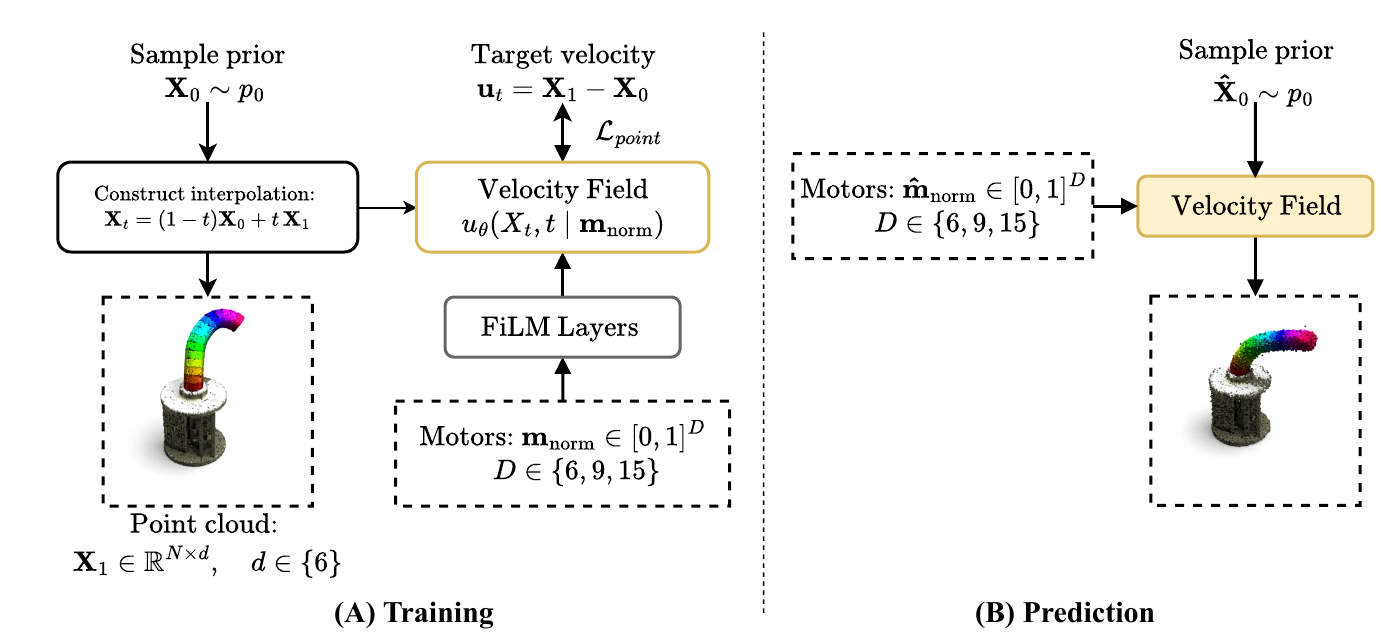}
    \caption{Action conditioned flow matching framework. (A) \textbf{Training:} sample Gaussian prior $X_0$, select a ground truth settled point cloud $X_1$ with condition $c$ (motor command, optionally augmented with payload), construct the interpolation state $X_t$, and supervise the velocity field $u_\theta$ with the target transport direction. Conditions are injected via FiLM layers. (B) \textbf{Prediction:} integrate the learned ODE from noise to generate the predicted steady state point cloud $\hat{X}^{\star}$ conditioned on a query $\hat{c}$. The integration variable is flow time, not physical deformation time.}
    \label{fig:method_pipeline}
    \vspace{-0.7em}
\end{figure*}

All point clouds are represented in a fixed robot/world frame.
During training, we use an origin anchored global scaling
\begin{equation}
\tilde{X}=X/s,
\label{eq:point_normalization}
\end{equation}
where $s$ is a dataset level scale factor and no translation is applied, preserving the shared base coordinate system.
Motor commands and optional payload values are normalized per dimension using dataset statistics, e.g.,
\begin{equation}
\tilde{m}_{j}=\frac{m_j-(m_{\min})_j}{(m_{\max})_j-(m_{\min})_j},\qquad j=1,\ldots,D,
\label{eq:motor_normalization}
\end{equation}
and payload values are normalized analogously.
At inference, metric predictions are recovered by multiplying predicted XYZ coordinates by $s$.
Figure~\ref{fig:method_pipeline} summarizes the training and prediction pipeline.

\subsection{TDCR Hardware Platform and RGB-D Observation}
\label{subsec:hardware}
The learning algorithm only requires pairs of actuation states and observed point clouds, but the physical platform defines the real hardware action and observation spaces.
We use a lightweight 3D printed TDCR platform instantiated as 2- and 3-module real robots.
Each module is driven by three tendons arranged at $120^\circ$ around a PLA structural backbone, so an $S$-module robot uses $3S$ motors.
The robot is configurable in module count at fabrication/assembly time, but a finished robot is not intended for rapid hot swapping of modules.
The DYNAMIXEL implementation used in the real experiments has robot masses of $582.33$\,g (2 modules) and $767.96$\,g (3 modules).
Detailed tendon routing, actuator options, BOM costs, spring thickness choices, and manufacturing notes are provided in Appendix~\ref{supp:hardware}.

\begin{figure*}[t]
    \centering
    \includegraphics[width=0.92\textwidth]{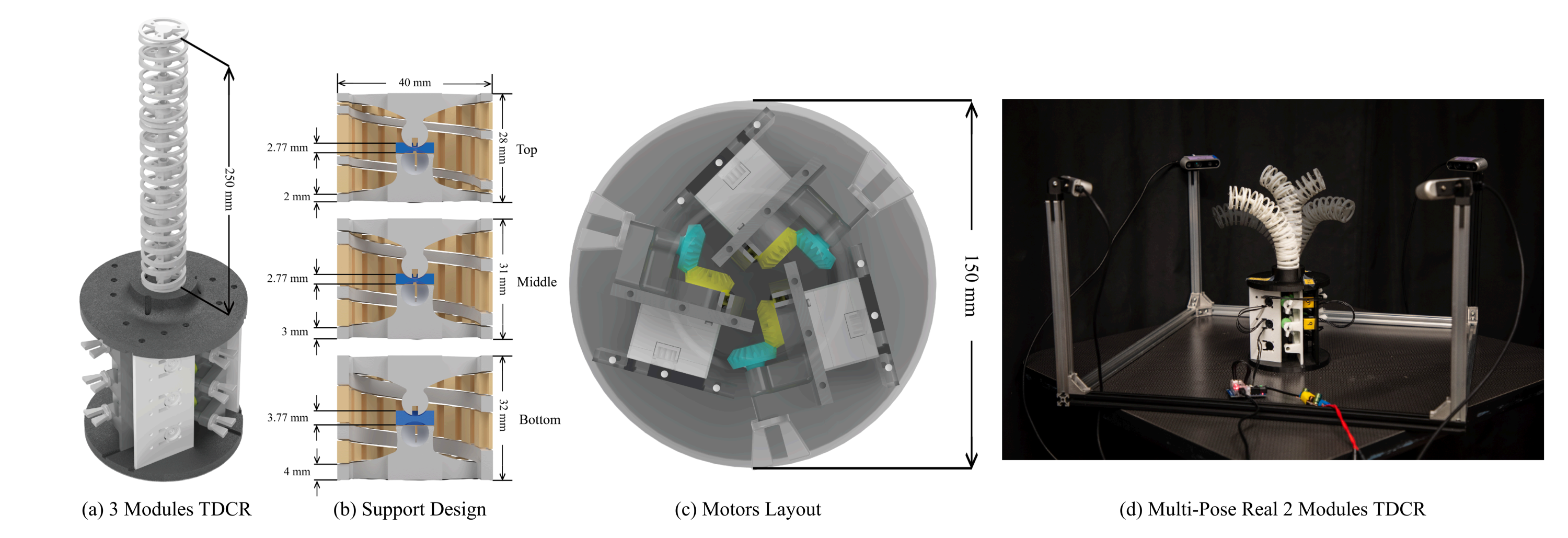}
    \caption{TDCR hardware platform and real hardware data capture. (a) Three module CAD rendering. (b) Support design for single pass 3D printing. (c) Actuation base and motor layout. (d) Real two module TDCR during RGB-D data collection.}
    \label{fig:tdcr_hardware}
    \vspace{-1.2em}
\end{figure*}

For real hardware self modeling data, four calibrated Intel RealSense D435 RGB-D cameras surround the robot.
For each settled command, we back project synchronized RGB-D views using camera intrinsics, transform them into a shared world frame with calibrated camera to camera extrinsics, and fuse them into a colored point cloud.
The resulting point clouds are voxel downsampled and randomly sampled or padded to a fixed point count for learning.
This representation gives the model direct supervision on full body 3D geometry rather than on a sparse backbone state or a hand designed low dimensional shape parameterization.

\subsection{Action Conditioned Flow Matching in Point Space}
\label{subsec:flowmatching}
We learn a conditional velocity field that transports a simple point prior to the distribution of settled TDCR shapes for a queried condition~\cite{chen2018neural,lipman2023flowmatching}.
Let $X_0\sim p_0=\mathcal{N}(0,\sigma^2 I)$ be a Gaussian point cloud with the same shape as a data sample $X_1$.
For a randomly sampled flow time $t\in[0,1]$, a straight interpolation path is
\begin{equation}
X_t=(1-t)X_0+tX_1,
\label{eq:linear_path_tdcr}
\end{equation}
with pairwise transport velocity $u_t=X_1-X_0$.
The learning target is the marginal conditional velocity field
\begin{equation}
u^\star(X,t\mid c)=\mathbb{E}\!\left[X_1-X_0\,\middle|\,X_t=X,\,t,\,c\right].
\label{eq:marginal_velocity_tdcr}
\end{equation}
The continuous variable $t$ in Eqs.~\eqref{eq:linear_path_tdcr}--\eqref{eq:marginal_velocity_tdcr} is the generative flow matching transport time; it is not the physical time of TDCR deformation.

We parameterize the velocity field with a neural network $u_\theta(X_t,t\mid c)$ and optimize
\begin{equation}
\mathcal{L}_{\mathrm{FM}}
=
\mathbb{E}_{t,X_0,X_1,c}
\left[
\left\|u_\theta(X_t,t\mid c)-(X_1-X_0)\right\|_F^2
\right],
\label{eq:fm_loss_tdcr}
\end{equation}
where $\|\cdot\|_F$ is the Frobenius norm over the full $N\times d$ point cloud residual matrix.
We sample $t\sim\mathrm{Beta}(\alpha,1)$ with $\alpha>1$ to emphasize later interpolation states near the data distribution.
For colored point clouds, the loss is split into geometry and color terms,
\begin{equation}
\mathcal{L}_{\mathrm{FM}}=\mathcal{L}_{\mathrm{xyz}}+\lambda_{\mathrm{rgb}}\mathcal{L}_{\mathrm{rgb}},
\label{eq:rgb_loss_tdcr}
\end{equation}
with $\lambda_{\mathrm{rgb}}=0.05$ in our experiments.
After training, a prediction for a query condition $\hat{c}$ is generated by sampling $\hat{X}_0\sim p_0$ and integrating
\begin{equation}
\dot{X}_t=u_\theta(X_t,t\mid \hat{c}),\qquad X_0=\hat{X}_0,
\label{eq:ode_tdcr}
\end{equation}
from $t=0$ to $t=1$.
The endpoint $X_{t=1}$ of this generative ODE is the predicted settled point cloud.

\subsection{Condition Injection and Velocity Network Architectures}
\label{subsec:architecture}
We evaluate two interchangeable architectures for $u_\theta$, both permutation equivariant with respect to point ordering.
Throughout this subsection, ``velocity network'' refers to the neural architecture; the mechanical component of the robot is referred to as the structural backbone.

\paragraph{Time and action embeddings}
We embed flow time $t$ using sinusoidal features and project the condition vector $c$ with a linear layer.
The two embeddings are fused additively,
\begin{equation}
e(t,c)=\phi_t(t)+\phi_c(c),
\label{eq:embedding_fusion}
\end{equation}
and injected into intermediate features using FiLM modulation~\cite{perez2018film}:
\begin{equation}
\mathrm{FiLM}(h;e)=(1+\gamma(e))\odot\mathrm{LN}(h)+\beta(e),
\label{eq:film}
\end{equation}
where $\gamma(\cdot)$ and $\beta(\cdot)$ are learned affine heads, $\mathrm{LN}$ is LayerNorm, and $\odot$ denotes element wise multiplication.

\paragraph{MLP velocity network}
The lightweight MLP architecture processes each point independently with shared weights.
For each point $x_i\in\mathbb{R}^{d}$, we concatenate its current coordinates/features with the global embedding and predict a per point velocity:
\begin{equation}
u_\theta(X_t,t\mid c)=\{f_\theta([x_i\,\|\,e(t,c)])\}_{i=1}^{N}.
\label{eq:mlp_velocity}
\end{equation}
Residual MLP blocks with FiLM modulation provide an efficient $O(N)$ architecture for large point clouds.

\paragraph{Hybrid PVConv velocity network}
To model local geometric context, the Hybrid architecture augments the per point head with a multi scale point voxel context module.
A PVConv pyramid~\cite{liu2019pvcnn} extracts per point context features
\begin{equation}
C_{\mathrm{pv}}=\mathrm{ContextNet}(X_t,t,c)\in\mathbb{R}^{N\times d_c},
\label{eq:contextnet}
\end{equation}
which are concatenated with each point and the global embedding before the FiLM MLP head predicts velocities.
Because the early part of the generative trajectory starts from unstructured Gaussian noise, we use a time gated blend between global and local context,
\begin{equation}
C(t)=\alpha(t)C_{\mathrm{pv}}+(1-\alpha(t))C_{\mathrm{global}},\quad
\alpha(t)=\sigma(k(t-\tau)),
\label{eq:time_gate}
\end{equation}
where $\sigma(\cdot)$ is the sigmoid and $k,\tau$ control the transition from global conditioning to local geometric context.

\subsection{Training and Shape Prediction Algorithms}
\label{subsec:training_sampling}
Algorithms~\ref{alg:tdcr-training} and~\ref{alg:tdcr-sampling} summarize the implementation used for training and inference.
During training, we draw a settled point cloud and condition vector, sample a Gaussian point cloud, construct the interpolation state, and regress the conditional transport velocity.
During prediction, we integrate Eq.~\eqref{eq:ode_tdcr} with Heun's method (improved Euler/RK2) and use EMA model weights.

\begin{algorithm}[t]
\footnotesize
\caption{TDCR Training (Action-Conditioned Flow Matching)}
\label{alg:tdcr-training}
\begin{algorithmic}[1]
\Require Dataset $\mathcal{D}=\{(X_1^{i},c^{i})\}_{i=1}^{K}$, where $c=\tilde{m}$ or $[\tilde{m}\,\|\,\tilde{p}]$; prior $p_0=\mathcal{N}(0,\sigma^2 I)$; velocity field $u_\theta$; optimizer $\textsc{Opt}$.
\For{step $=1$ to $T$}
    \State Sample minibatch $(X_1,c) \sim \mathcal{D}$
    \State Sample $X_0 \sim p_0$ \Comment{same $N\times d$ shape as $X_1$}
    \State Sample $t \sim \mathrm{Beta}(\alpha,1)$
    \State Construct $X_t \gets (1-t)X_0+tX_1$
    \State Target velocity $u_t \gets X_1-X_0$
    \State Predict $\hat{u} \gets u_\theta(X_t,t\mid c)$ \Comment{$\hat{u}\in\mathbb{R}^{N\times d}$}
    \State $\mathcal{L}_{\mathrm{xyz}} \gets \|\hat{u}_{\mathrm{xyz}}-u_{t,\mathrm{xyz}}\|_F^2$
    \If{$d=6$}
        \State $\mathcal{L}_{\mathrm{rgb}} \gets \|\hat{u}_{\mathrm{rgb}}-u_{t,\mathrm{rgb}}\|_F^2$
        \State $\mathcal{L} \gets \mathcal{L}_{\mathrm{xyz}}+\lambda_{\mathrm{rgb}}\mathcal{L}_{\mathrm{rgb}}$
    \Else
        \State $\mathcal{L} \gets \mathcal{L}_{\mathrm{xyz}}$
    \EndIf
    \State Update $\theta \leftarrow \textsc{Opt.step}(\nabla_\theta \mathcal{L})$
    \State Update EMA weights
\EndFor
\end{algorithmic}
\end{algorithm}

\begin{algorithm}[t]
\footnotesize
\caption{TDCR Shape Prediction (Heun RK2 Sampling)}
\label{alg:tdcr-sampling}
\begin{algorithmic}[1]
\Require Trained velocity field $u_\theta$ with EMA weights; query condition $\hat{c}$; prior $p_0$; number of steps $S$; step size $h\gets 1/S$.
\State Sample $X \sim p_0$ \Comment{$X\in\mathbb{R}^{N\times d}$}
\For{$k=0$ \textbf{to} $S-1$}
    \State $t_0 \gets kh$
    \State $v_1 \gets u_\theta(X,t_0\mid \hat{c})$
    \State $\bar{X} \gets X+h\,v_1$
    \State $t_1 \gets (k+1)h$
    \State $v_2 \gets u_\theta(\bar{X},t_1\mid \hat{c})$
    \State $X \gets X+\frac{h}{2}(v_1+v_2)$
\EndFor
\State \Return $\hat{X}^{\star}\gets X$ \Comment{multiply XYZ by $s$ to recover metric scale}
\end{algorithmic}
\end{algorithm}

\section{Experiments}
\label{sec:experiments}

We evaluate action conditioned flow matching on simulated and real TDCR self modeling tasks.
Each experiment predicts the full settled 3D robot shape, represented as a point cloud, from a commanded actuation state.
Unless otherwise specified, all methods are trained and evaluated per robot morphology and dataset split.
The primary setting is unloaded/free space quasi static prediction; Sec.~\ref{subsec:payload_extension} evaluates a payload conditioned simulation extension.


\begin{figure}[!b]
    \centering
    \includegraphics[width=\columnwidth]{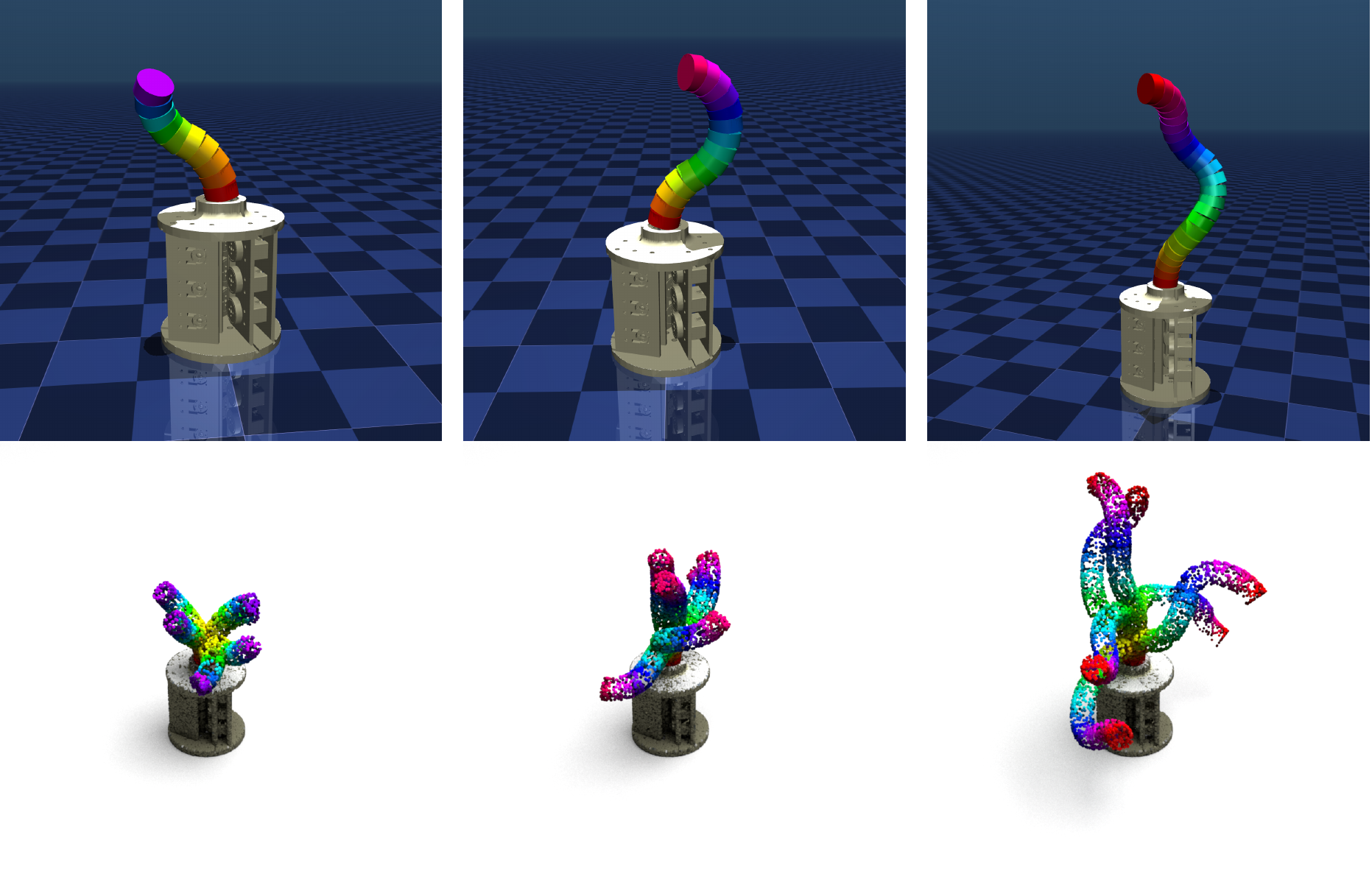}
    \caption{MuJoCo renderings of our tendon driven continuum robot (TDCR) simulators for 2/3/5 modules (with base). Colors indicate the disc index along the backbone. Top row: single representative poses. Bottom row: multiple poses overlaid, illustrating the diversity of configurations in our dataset. The same simulated scene and camera setup is used to generate the RGB-D observations from multiple cameras.}
    \label{fig:mujoco}
    \vspace{-0.7em}
\end{figure}
\begin{figure*}[!t]
    \centering
    \includegraphics[width=0.98\textwidth]{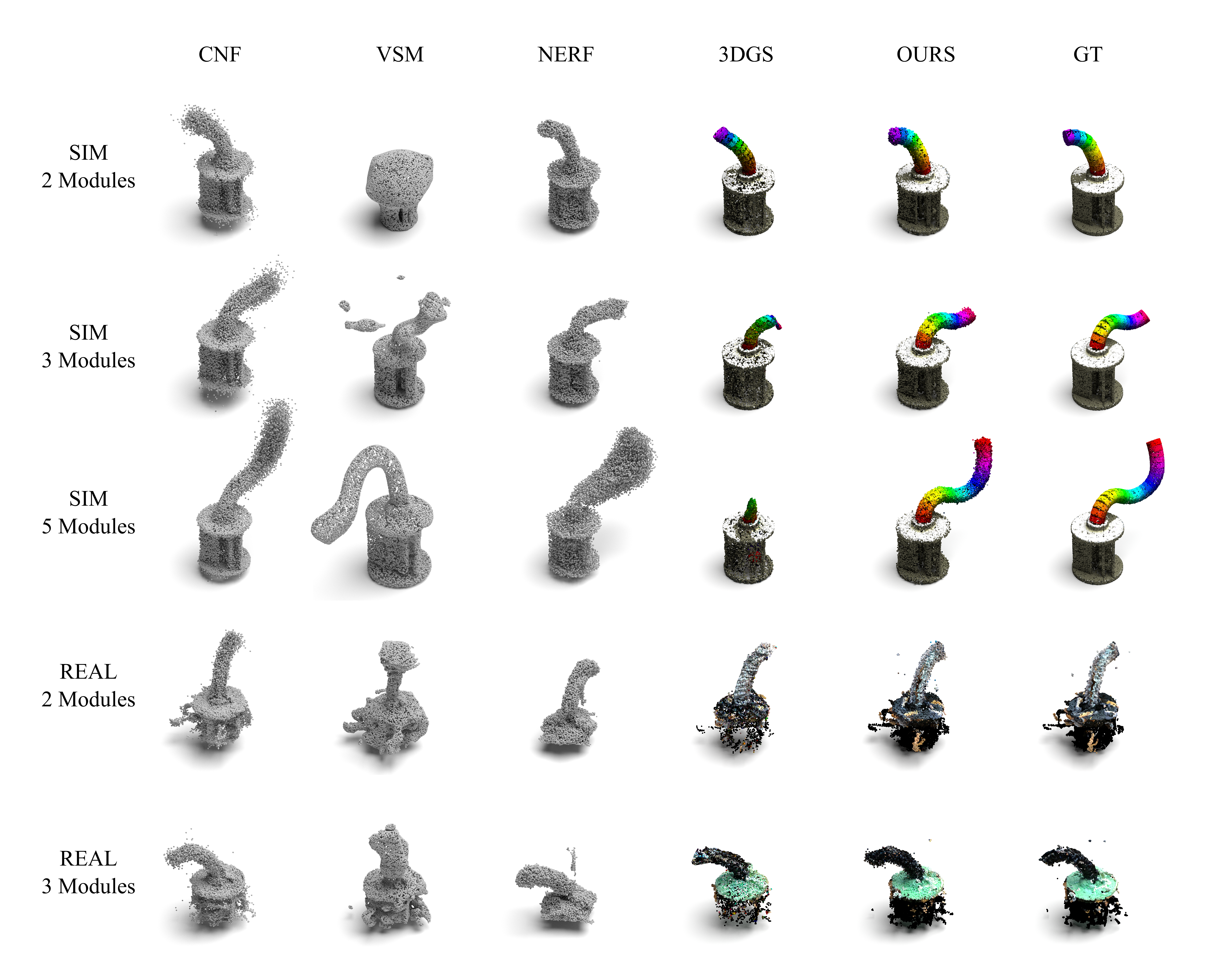}
    \caption{Qualitative comparison on simulated and real TDCR datasets. Rows show simulated 2-/3-/5-module with base settings and real 2-/3-module hardware. Columns show CNF (PointFlow), VSM, NeRF (FFKSM), 3DGS, our method, and the ground truth (GT). For our method, we visualize the best performing velocity network architecture (MLP or Hybrid) selected by validation CD.}
    \label{fig:qual_sim}
    \vspace{-0.8em}
\end{figure*}

\subsection{Datasets and Metrics}
\label{sec:exp_data_metrics}

\paragraph{Metrics}
Following prior point based self modeling work, we report the squared L2 Chamfer Distance (CD) and Earth Mover's Distance (EMD) between the predicted point cloud $\hat{X}$ and the ground truth point cloud $X$.
Let $P,Q\subset\mathbb{R}^3$ be two sets of points. The squared L2 Chamfer Distance is
\begin{equation}
\mathrm{CD}(P,Q)=
\frac{1}{|P|}\sum_{p\in P}\min_{q\in Q}\|p-q\|_2^2
+
\frac{1}{|Q|}\sum_{q\in Q}\min_{p\in P}\|q-p\|_2^2 ,
\label{eq:cd}
\end{equation}
and EMD is the minimum transport cost under a bijection $\pi$,
\begin{equation}
\mathrm{EMD}(P,Q)
=
\min_{\pi\in\Pi}\;
\frac{1}{|P|}\sum_{i=1}^{|P|}\|p_i - q_{\pi(i)}\|_2 ,
\label{eq:emd}
\end{equation}
where $\Pi$ is the set of all permutations.
For fair comparison, we compute both CD and EMD on point clouds of equal size by uniformly sampling $N_{\text{eval}}{=}10{,}000$ points from both $P$ and $Q$.

\paragraph{Normalization and evaluation scale}
All models are trained on globally normalized point clouds and normalized condition vectors (Sec.~\ref{sec:method}).
For reporting CD/EMD, we invert the point normalization and compute metrics in the original metric scale:
\begin{equation}
X_{\text{metric}} = \tilde{X}\cdot s ,
\end{equation}
where $s$ is the scale factor for the entire dataset under origin anchored scaling without translation.

\paragraph{Simulation datasets}
We build a family of tendon driven continuum robot simulators in MuJoCo.
Each module is approximated as a chain of rigid disks connected by ball joints, while tendons are implemented using MuJoCo spatial tendons that pass through fixed sites on each disk.
Actuators control tendon rest lengths, producing bending and twisting responses similar to physical TDCRs.
This modeling choice is inspired by recent musculoskeletal simulation practices that rely on tendon like elements for complex deformation~\cite{ostrichrl}.

We created six standard simulation datasets: \textbf{2m}, \textbf{3m}, and \textbf{5m} (2/3/5 modules) $\times$ \textbf{without} base and \textbf{with} base.
The \emph{with base} variant includes the rigid base in the rendered observations, whereas the \emph{no base} variant excludes it via XML visibility/group settings so that the point cloud contains only the deformable body.
The no base setting is an analysis only ablation: with a fixed point budget, removing the rigid base increases sampling density on the deformable body.
Because the compared methods use different point, implicit, and rendering objectives, the no base and with base settings can show different trends; we use this ablation to analyze sensitivity to rigid scene geometry rather than to argue that removing the base always improves prediction.

For each dataset, we uniformly sample motor commands within actuator limits and let the simulator relax to a static configuration before capture.
We render RGB-D observations from a ring of fixed cameras around the robot, apply segmentation masks to keep foreground geometry, and back project depth to colored point clouds.
Each sample is voxel downsampled and then randomly sampled or padded to $N_{\text{train}}{=}20{,}000$ points for training.
Dataset sizes are $5{,}000$ motor configurations for all 2m/3m settings and $10{,}000$ for all 5m settings, with an $80\%/10\%/10\%$ train/val/test split.
Motor dimensions are $D=6$ (2 modules), $D=9$ (3 modules), and $D=15$ (5 modules).
Figure~\ref{fig:mujoco} shows example MuJoCo renderings of the three TDCR morphologies.

\paragraph{Payload conditioned simulation datasets}
For the payload extension, we use the with base simulation scenes and augment the condition from $\tilde{m}$ to $[\tilde{m}\,\|\,\tilde{p}]$, where $p$ is a scalar tip payload mass.
For each sample, the simulator first settles under the sampled motor command, then applies a gravity aligned tip load $f_p=pg$ to the distal body, relaxes again, and captures the loaded steady state point cloud.
Payload masses are sampled from $[0,0.030]$ kg, with a $10\%$ probability of exactly zero load.
We collect $5{,}000$ motor and payload configurations for each loaded 2m/3m/5m with base morphology and use the same $80\%/10\%/10\%$ split.
This setting evaluates payload conditioned quasi static shape prediction, not arbitrary contact, torque, or dynamic loading.

\paragraph{Real hardware datasets}
We collect real hardware datasets with physical 2-module and 3-module TDCRs (Sec.~\ref{subsec:hardware}).
Motor commands are generated via a joystick to motor mapping, executed to a settled configuration, and captured with four Intel RealSense D435 cameras.
We fuse calibrated RGB-D frames from multiple cameras into a colored point cloud in a shared world frame, then apply the same voxel downsampling, $N_{\text{train}}{=}20{,}000$ point sampling, and normalization as in simulation.

\begin{table*}[!t]
\centering
\small
\setlength{\tabcolsep}{4pt}
\renewcommand{\arraystretch}{1.08}
\caption{Simulation steady state shape prediction on TDCR datasets. Lower is better. We report $\mathrm{CD}\times 10^{4}$ and $\mathrm{EMD}\times 10^{3}$ (metric scale, $N_{\text{eval}}{=}10{,}000$ points). NB/WB denote no base/with base observations.}
\label{tab:sim_results}
\resizebox{\textwidth}{!}{%
\begin{tabular}{l|cccccc|cccccc}
\hline
 & \multicolumn{6}{c|}{$\mathrm{CD}\times 10^{4}\;\downarrow$} & \multicolumn{6}{c}{$\mathrm{EMD}\times 10^{3}\;\downarrow$} \\
Method
& 2m-NB & 2m-WB & 3m-NB & 3m-WB & 5m-NB & 5m-WB
& 2m-NB & 2m-WB & 3m-NB & 3m-WB & 5m-NB & 5m-WB \\
\hline
VSM~\cite{vsm} & 14.773 & 10.065 & 92.509 & 32.494 & 105.017 & 12.827 & 5.795 & 4.881 & 19.990 & 7.386 & 19.470 & 7.112 \\
PointFlow~\cite{pointflow} & 0.359 & 0.461 & 2.172 & 1.013 & 19.790 & 6.788 & 0.598 & 1.041 & 1.362 & 1.203 & 5.453 & 3.227 \\
NeRF~\cite{nerf_selfsim} & 0.341 & 0.953 & 0.650 & 0.974 & 6.552 & 4.309 & 1.009 & 2.263 & 2.958 & 2.035 & 9.753 & 15.854 \\
3DGS~\cite{gs_selfmodel} & 0.447 & 0.373 & 1.053 & 1.205 & 19.201 & 63.781 & 0.461 & 3.779 & 0.847 & 4.837 & 4.849 & 14.652 \\
\hline
Ours (MLP) & 0.100 & 0.147 & 0.158 & 0.194 & 0.260 & 0.260 & 0.230 & 0.587 & 0.349 & 0.672 & 0.870 & 1.110 \\
Ours (Hybrid) & \textbf{0.088} & \textbf{0.135} & \textbf{0.136} & \textbf{0.164} & \textbf{0.253} & \textbf{0.239} & \textbf{0.204} & \textbf{0.516} & \textbf{0.329} & \textbf{0.578} & \textbf{0.832} & \textbf{1.059} \\
\hline
\end{tabular}}
\vspace{0.35em}

\begin{minipage}[t]{0.47\textwidth}
\centering
\footnotesize
\setlength{\tabcolsep}{3pt}
\renewcommand{\arraystretch}{1.05}
\caption{Real hardware steady state shape prediction. Lower is better.}
\label{tab:real_results}
\resizebox{\columnwidth}{!}{%
\begin{tabular}{l|cc|cc}
\hline
 & \multicolumn{2}{c|}{$\mathrm{CD}\times 10^{4}\;\downarrow$} & \multicolumn{2}{c}{$\mathrm{EMD}\times 10^{3}\;\downarrow$} \\
Method & Real-2m & Real-3m & Real-2m & Real-3m \\
\hline
VSM~\cite{vsm} & 2.287 & 1.525 & 4.426 & 5.363 \\
PointFlow~\cite{pointflow} & 0.340 & 0.266 & 0.828 & 0.672 \\
NeRF~\cite{nerf_selfsim} & 9.569 & 10.377 & 11.111 & 9.251 \\
3DGS~\cite{gs_selfmodel} & 1.030 & 0.809 & 12.592 & 14.952 \\
\hline
Ours (MLP) & \textbf{0.226} & 0.215 & \textbf{0.522} & \textbf{0.507} \\
Ours (Hybrid) & 0.234 & \textbf{0.212} & 0.575 & 0.548 \\
\hline
\end{tabular}}
\end{minipage}
\hfill
\begin{minipage}[t]{0.49\textwidth}
\centering
\footnotesize
\setlength{\tabcolsep}{3.5pt}
\renewcommand{\arraystretch}{1.05}
\caption{Payload conditioned simulation extension. The condition is augmented from $\tilde{m}$ to $[\tilde{m}\,\|\,\tilde{p}]$, where $p$ is a scalar tip payload mass. Lower is better.}
\label{tab:payload_results}
\resizebox{\columnwidth}{!}{%
\begin{tabular}{l|ccc|ccc}
\hline
 & \multicolumn{3}{c|}{$\mathrm{CD}\times 10^{4}\;\downarrow$} & \multicolumn{3}{c}{$\mathrm{EMD}\times 10^{3}\;\downarrow$} \\
Method & 2m & 3m & 5m & 2m & 3m & 5m \\
\hline
Ours (MLP) & 0.393 & 0.386 & \textbf{0.741} & 0.992 & 0.826 & \textbf{1.272} \\
Ours (Hybrid) & \textbf{0.340} & \textbf{0.378} & 1.780 & \textbf{0.709} & \textbf{0.701} & 1.290 \\
\hline
\end{tabular}}
\end{minipage}
\vspace{-0.8em}
\end{table*}

\subsection{Baselines}
\label{sec:exp_baselines}

We compare against representative robot self modeling and 3D generative baselines, spanning point based and view based formulations.
All baselines are trained on the same train split as ours and evaluated on the same test split using the point cloud metrics in Sec.~\ref{sec:exp_data_metrics}.

\paragraph{VSM}
Visual Self-Modeling (VSM)~\cite{vsm} learns a robot self model from \texttt{xyzn} observations.
We follow the official data interface, estimate stable normals for each TDCR point cloud, and convert VSM outputs back to point clouds for CD/EMD evaluation.

\paragraph{CNF / PointFlow}
We include a conditional continuous normalizing flow baseline based on PointFlow~\cite{pointflow}.
The flow is conditioned on the normalized motor vector and trained on the same splits.
We use XYZ only point clouds and keep the number of generated points identical to our evaluation protocol.

\paragraph{NeRF self simulation (FFKSM)}
We adapt the self simulation method in \emph{Teaching Robots to Build Simulations of Themselves}~\cite{nerf_selfsim}, which introduces a Free Form Kinematic Self Model (FFKSM) inspired by NeRF style reasoning.
FFKSM is a query based implicit model that predicts occupancy/visibility of 3D query points conditioned on robot configuration and is trained from binary silhouettes.
We train it with $100\times100$ masks following the original protocol and extract point clouds by densely querying the workspace and thresholding predicted occupancy.

\paragraph{3D Gaussian Splatting (3DGS)}
We include the articulated 3D Gaussian splatting self model from \emph{Learning High Fidelity Robot Self Model with Articulated 3D Gaussian Splatting}~\cite{gs_selfmodel}.
This method reconstructs a robot with 3D Gaussians and learns a neural bone/LBS style deformation model conditioned on robot configurations.
We train 3DGS with RGB and mask supervision from multiple camera views at $1280\times720$ resolution and evaluate by fusing rendered depth from calibrated viewpoints into an XYZ point cloud.

\paragraph{Mask generation for view based baselines}
Both NeRF (FFKSM) and 3DGS require foreground masks.
For simulation, MuJoCo provides near perfect instance masks.
For real hardware captures, masks derived from depth thresholding are unreliable; we therefore use SAM~2~\cite{ravi2024sam2} to generate masks by manually annotating a small number of example masks per view and propagating to the remaining frames.

\subsection{Implementation Details}
\label{sec:exp_impl}

\paragraph{Ours}
We train our action conditioned flow matching models for $500$ epochs.
We sample flow time with $t\!\sim\!\mathrm{Beta}(\alpha,1)$ using $\alpha{=}3.0$ and sample the point prior as $X_0\!\sim\!\mathcal{N}(0,\sigma^2 I)$ with $\sigma{=}0.5$.
For RGB point clouds, we use the color loss weight $\lambda_{\text{rgb}}{=}0.05$.
We use Heun RK2 sampling with $S{=}100$ steps for both the MLP and Hybrid velocity networks.
For payload conditioned models, we use the same training procedure but set the condition mode to motor plus payload, i.e., $c=[\tilde{m}\,\|\,\tilde{p}]$.
Unless stated otherwise, we report results using EMA weights selected by validation CD.

\paragraph{Training and evaluation protocol}
All methods use the same train/val/test split within each dataset.
We train with $N_{\text{train}}{=}20{,}000$ points per sample from the normalized H5 datasets.
For evaluation, we denormalize predictions and ground truth to metric scale and uniformly sample $N_{\text{eval}}{=}10{,}000$ points from each before computing CD and EMD.

\subsection{Quantitative Results}
\label{sec:exp_quant}

Table~\ref{tab:sim_results} reports standard simulation results, and Table~\ref{tab:real_results} reports real hardware results.
For readability, we report $\mathrm{CD}\times 10^{4}$ and $\mathrm{EMD}\times 10^{3}$, computed on $N_{\text{eval}}{=}10{,}000$ points in metric scale.

Across the six standard simulation settings, our Hybrid model achieves the lowest CD and EMD.
Relative to the strongest non ours baseline in each setting, it reduces CD by approximately $64$--$96\%$ and EMD by approximately $50$--$83\%$.
On real data, the best of our two velocity networks reduces CD relative to PointFlow by $33.5\%$ on Real-2m and $20.3\%$ on Real-3m, and reduces EMD by $36.9\%$ and $24.6\%$, respectively.
These results indicate that direct point cloud flow matching is effective for predicting full body TDCR geometry from test actuation states.

\subsection{Qualitative Results}
\label{sec:exp_qual}

Figure~\ref{fig:qual_sim} compares representative predictions on simulated and real datasets.
PointFlow captures global pose but can produce diffuse samples near the distal body.
View based baselines are sensitive to mask quality and thin body geometry: NeRF (FFKSM) degrades in complex 5-module scenes, and 3DGS may reconstruct the rigid base more reliably than the thin continuum body.
Our method maintains cleaner tip and body geometry across morphologies.

\subsection{Payload Conditioned Simulation Extension}
\label{subsec:payload_extension}

\begin{figure}[!t]
    \centering
    \includegraphics[width=0.98\columnwidth]{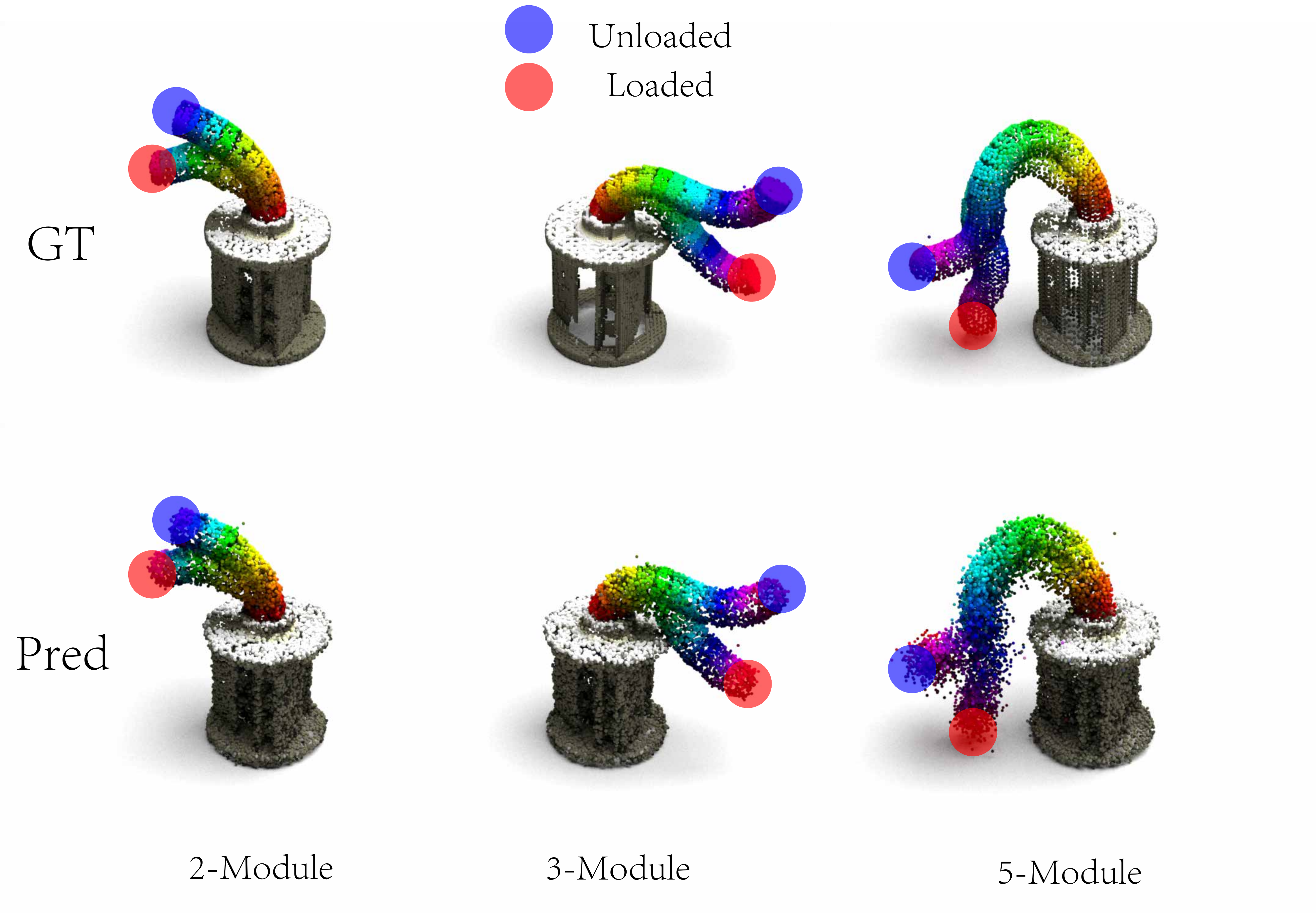}
    \caption{Payload conditioned prediction in simulation. Blue/red markers show unloaded and maximum load tip states; rows show ground truth and prediction.}
    \label{fig:payload_loaded}
    \vspace{-0.8em}
\end{figure}

Table~\ref{tab:payload_results} and Fig.~\ref{fig:payload_loaded} evaluate the payload conditioned extension.
The model is trained on loaded steady state point clouds with condition $c=[\tilde{m}\,\|\,\tilde{p}]$, where $p$ is the scalar payload mass and the applied force is aligned with gravity.
Both velocity network variants remain accurate under payload conditioning for 2-/3-/5-module TDCRs, and the overlays show load induced tip deflection.
Hybrid is not uniformly better in this extension: it improves 2m/3m EMD but has a larger 5m CD than MLP.
We treat this as evidence that additional quasi static state variables can be incorporated when such data are available, not as a claim of general load dependent mechanics.

\subsection{Discussion}
\label{sec:exp_discussion}

\paragraph{Takeaways}
The standard simulation and real hardware results show that action conditioned flow matching can learn accurate quasi static TDCR self models for test commands within a fixed morphology and actuation range.
The dense point cloud objective directly matches the fused RGB-D observation space, while the payload extension shows that the same conditioning mechanism can include a scalar tip load variable in simulation.
Among baselines, CNF/PointFlow is the strongest point based competitor on real data but produces diffuse distal samples; NeRF/FFKSM and 3DGS are more sensitive to visibility, masks, and rigid base geometry.
CNF also requires substantially more wall clock training time on simulated with base datasets: 21.5/18.2/60.5 hours for 2m/3m/5m, versus 8.8/10.9/17.4 hours for our Hybrid model.

\paragraph{Scope and limitations}
Our experiments target observation level quasi static shape prediction, not transient dynamics; the flow matching ODE is a generative transport process rather than physical deformation time.
The real hardware experiments use 2- and 3-module TDCRs, whereas the 5-module and payload conditioned results are simulation only.
The payload experiment uses a scalar gravity aligned tip load and does not cover arbitrary contact, off axis torque, distributed load, unknown object geometry, or zero shot transfer across unseen continuum robot designs and materials.
Extending the framework to richer contact, distributed loads, and unseen TDCR designs is left for future work, and we view the current results as a foundation for these directions.

\def\tmplabel#1{[#1]}

\section{Conclusion}
\label{sec:conclusion}

We presented an action conditioned flow matching framework for quasi static self modeling of tendon driven continuum robots.
Given a normalized actuation state, and optionally a scalar tip payload condition in simulation, the model predicts the robot's settled 3D geometry as a dense point cloud.
Across MuJoCo 2-, 3-, and 5-module TDCR simulations and real 2- and 3-module hardware experiments, our method achieves lower Chamfer Distance and Earth Mover's Distance than representative point based and view based self modeling baselines.
The payload conditioned simulation extension further shows that the same conditional formulation can incorporate a gravity aligned tip payload magnitude when corresponding training data are available.
The method predicts steady state geometry rather than transient deformation trajectories, and our experiments do not claim universal mechanics modeling, arbitrary contact/load handling, or zero shot transfer across unseen continuum robot designs.
Future work will study real payload experiments, history- and contact aware conditioning, morphology conditioned transfer, and integration into closed loop planning and control.



\bibliographystyle{plainnat}
\bibliography{references}

\clearpage
\maketitlesupplementary

\begin{strip}
    \centering
    \includegraphics[width=\textwidth]{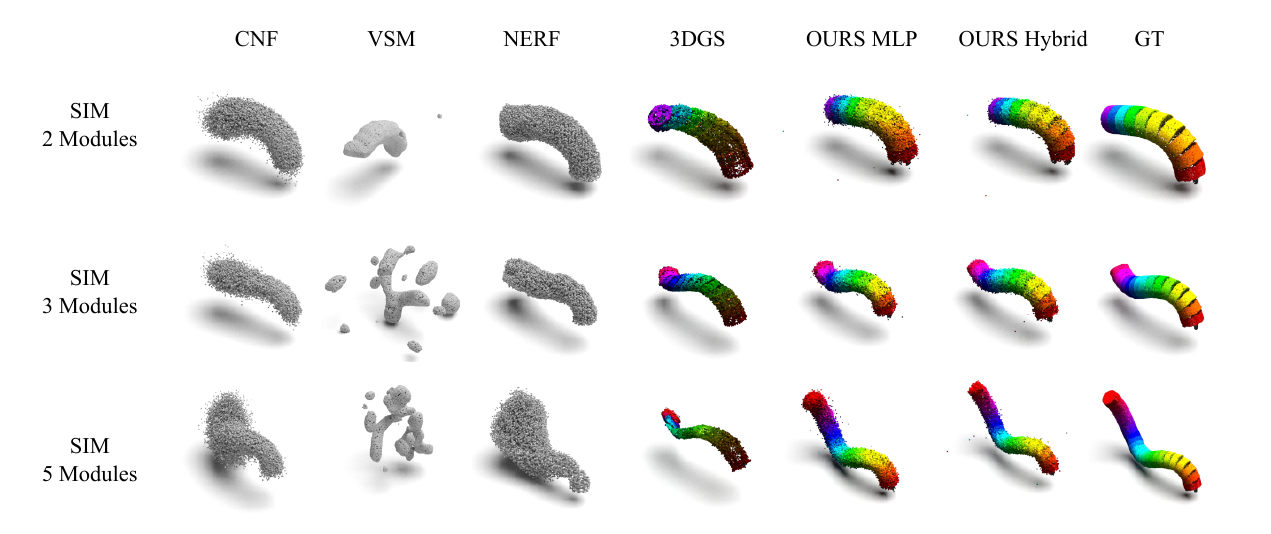}
    \captionof{figure}{Qualitative comparison on simulated \emph{no base} datasets (2/3/5 modules). Columns show CNF (PointFlow), VSM, NeRF (FFKSM), 3DGS, our method, and the ground truth (GT). For our method, we visualize both MLP and Hybrid velocity networks.}
    \label{fig:supp_qual_sim_nobase}
    \vspace{-0.5em}
\end{strip}

\section{Qualitative Comparison on \emph{no base} Datasets}
As shown in Fig.~\ref{fig:supp_qual_sim_nobase}, our method yields the most accurate and visually stable predictions across all \emph{no base} settings (2/3/5 modules).
The Hybrid velocity network produces cleaner reconstructions near the robot tip, with reduced jitter and noise.
VSM nearly fails to produce meaningful predictions in this regime, while NeRF self simulation (FFKSM) and 3D Gaussian Splatting (3DGS) degrade as the number of modules increases.
Among the baselines, CNF (PointFlow) is the strongest point based competitor.

\subsection{Manufacturing and Sensing System Details}
\label{supp:hardware}
This section provides the hardware details moved out of the main paper to satisfy the main text page limit.
The real experiments use the compact DYNAMIXEL implementation.
Table~\ref{tab:supp_hardware_summary} summarizes robot mass and approximate BOM costs; costs exclude cameras, power supply, and labor.

\begin{center}
\footnotesize
\captionof{table}{TDCR hardware variants and approximate BOM costs.}
\label{tab:supp_hardware_summary}
\begin{tabular}{lcccc}
\hline
Variant & Modules & Motors & Robot mass & BOM \\
\hline
DYNAMIXEL Real-2m & 2 & 6 & 582.33\,g & \$696.93 \\
DYNAMIXEL Real-3m & 3 & 9 & 767.96\,g & \$1006.93 \\
\hline
\end{tabular}
\end{center}

\paragraph{Print in place structure}
During printing, dedicated sacrificial support structures support the spring geometry and ball joint components.
The ball joints are printed in a disengaged state, with the ball and socket separated.
By adjusting the designed initial clearance between the ball and socket before assembly, we can tune the resulting module compliance once the joints are engaged.
All support components are designed to be removable after printing.
The structural backbone is fabricated as a single piece PLA print with repeating spring elements connected by ball joint interfaces.
Each module contains four spring units, and neighboring spring units use a $90^\circ$ helical phase shift to reduce nominal sideways bending at zero tension.
For the 3-module robot, the structural backbone spring thickness decreases from base to tip as 4/3/2\,mm, selected empirically across prototypes to reduce lower segment loading and nominal bending bias.

\paragraph{Tendon routing and module to motor assignment}
We use \texttt{nylon coated stainless steel bead stringing wire} as tendons.
Tendons are routed through guide holes in the structural backbone, with three tendons distributed at $120^\circ$ intervals around each module.
For the 2-module TDCR, the lower motor layer controls the lower structural backbone module and the upper motor layer controls the upper module.
For the 3-module TDCR, the three motor layers control the lower, middle, and upper modules, respectively.
Each tendon is anchored at the top of its target module, passes through the lower guide channel, and attaches to the corresponding motor spool for winding/unwinding.
A bevel gear interface on each spool enables manual tendon pretensioning at the home pose.
Figure~\ref{fig:supp_tendon_routing} visualizes the layer wise tendon assignment.

\begin{center}
    \includegraphics[width=0.95\columnwidth]{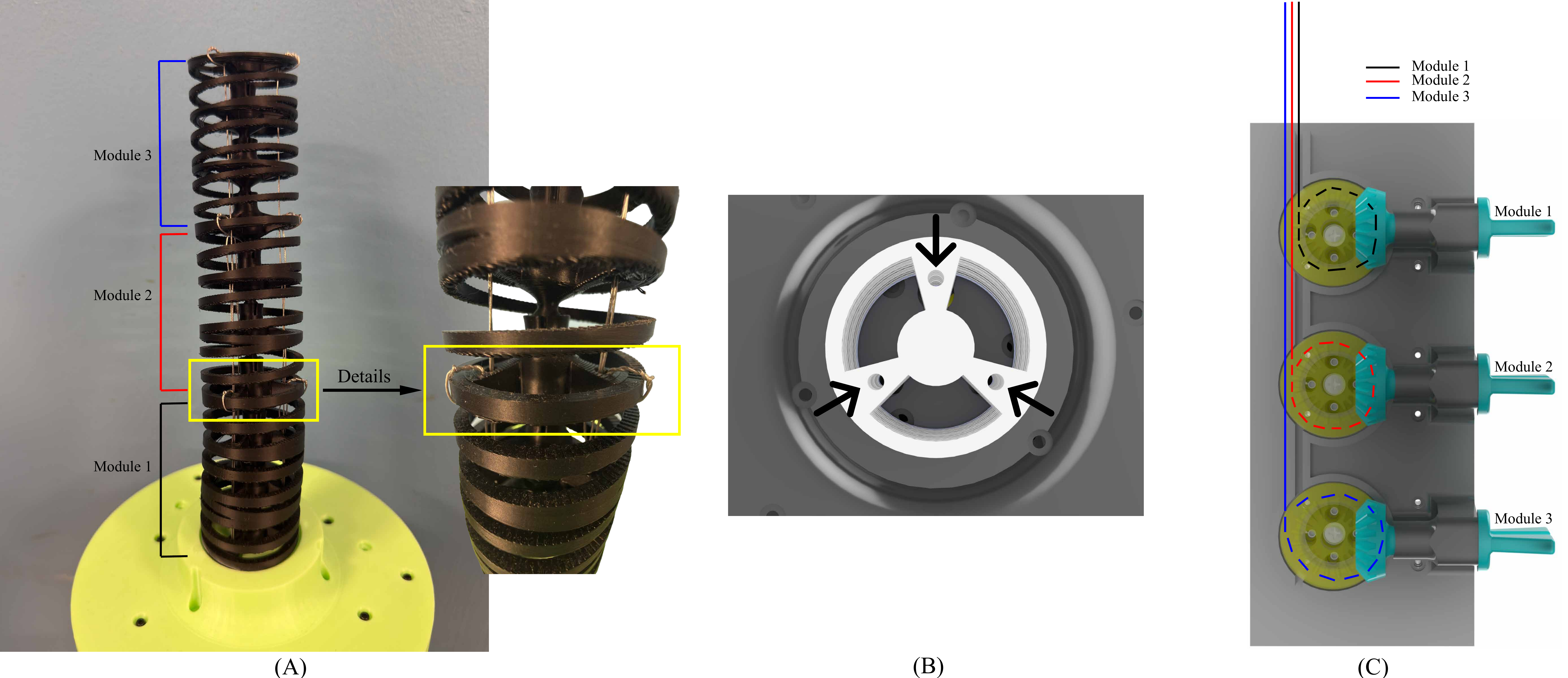}
    \captionof{figure}{Tendon routing overview. (A) Assembled 3-module robot and close-up. (B) Each module shares three tendon guide holes at $120^\circ$. (C) A tendon is anchored at the top of its assigned module, passes through the lower guide channel, and winds onto the spool in the corresponding motor layer.}
    \label{fig:supp_tendon_routing}
\end{center}

\paragraph{Real hardware sensing setup}
For real hardware data capture, we surround the TDCR with four \texttt{Intel RealSense D435} RGB-D cameras.
We estimate camera extrinsics via a camera to camera calibration procedure and use the factory depth-to-RGB calibration within each device to fuse synchronized depth and RGB images into colored point clouds.
We back project each RGB-D view using camera intrinsics, transform all views into a shared coordinate frame using calibrated extrinsics, and merge point clouds from multiple cameras to obtain the observation used for learning.

\subsection{Simulated Workspace Comparison}
\label{supp:workspace}
Figure~\ref{fig:supp_workspace} compares the reachable workspace of the simulated 2-, 3-, and 5-module TDCRs.
We show the YZ-plane projection of the reachable tip positions, together with the outer boundary of the sampled workspace.
As expected, increasing the number of modules expands both the lateral reach and the vertical span of the robot.
This figure helps explain why the higher-module settings are more challenging: the robot explores a larger shape space and exhibits larger overall deformation.

\begin{center}
    \includegraphics[width=0.98\columnwidth]{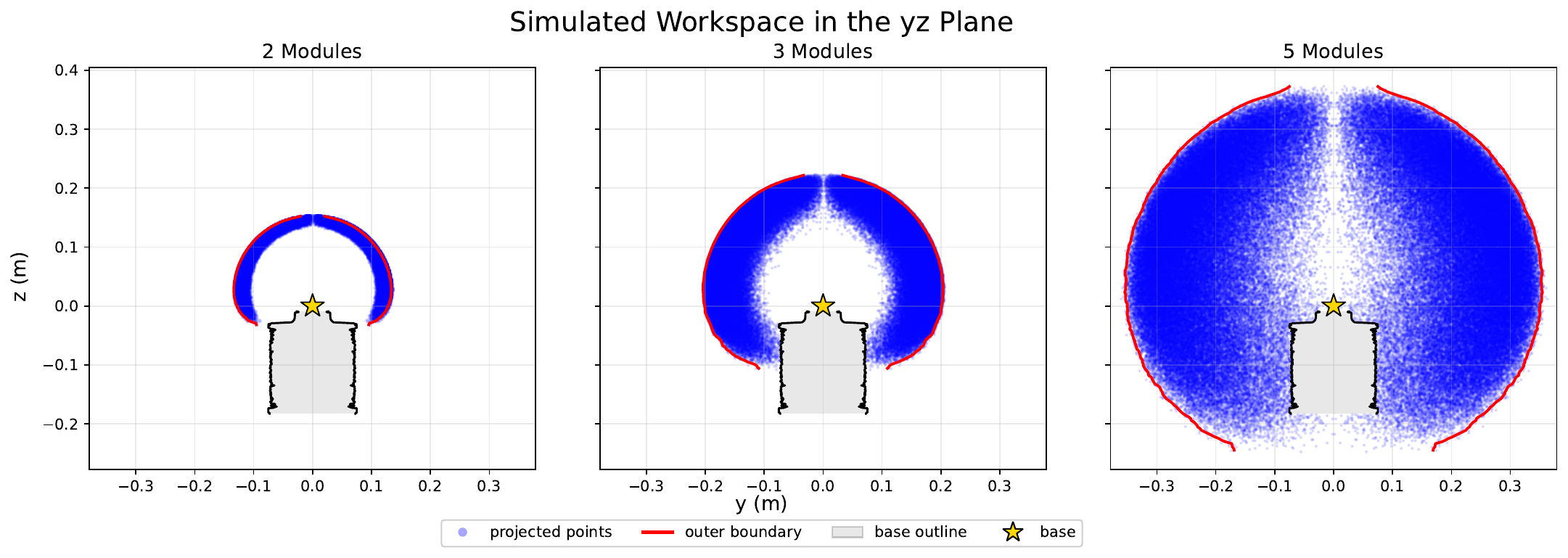}
    \captionof{figure}{YZ-plane workspace comparison for the simulated 2-, 3-, and 5-module TDCRs. Blue dots show sampled reachable tip positions, and the red curve outlines the outer workspace boundary.}
    \label{fig:supp_workspace}
\end{center}

\subsection{Open Source Release}
\label{supp:opensource}
For reproducibility, we have publicly released the full project at
\href{https://github.com/ruanjinchen/Continuum-Robot-Modeling-with-Action-Conditioned-Flow-Matching}
{\texttt{our GitHub repository}},
including:
(i) all code for training and inference,
(ii) trained checkpoints for all experiments,
(iii) the simulation models/assets used in our simulated datasets,
and (iv) the SolidWorks CAD model of the real hardware TDCR.

\end{document}